\documentclass[runningheads]{llncs}
\usepackage{graphicx}
\usepackage{amsmath,amssymb} 
\usepackage{color}
\usepackage[width=122mm,left=12mm,paperwidth=146mm,height=193mm,top=12mm,paperheight=217mm]{geometry}
\usepackage{booktabs} 
\usepackage{multirow}
\usepackage{subcaption}
\usepackage{wrapfig}
\usepackage[table]{xcolor} 
\usepackage{eso-pic}
\usepackage[export]{adjustbox}

\begin{document}
\pagestyle{headings}
\mainmatter
\def\ECCV16SubNumber{7690}  

\title{MorphoQuant: Modality-Aware Quantization for Omni-modal Large Language Models} 

\titlerunning{ }

\authorrunning{ }

\author{Yue Wu, Changyuan Wang, Zixuan Wang, Shilin Ma, Yansong Tang}
\institute{ }

\maketitle

\begin{abstract}
Conventional Post-Training Quantization (PTQ) methods struggle with 4-bit Omni-modal Large Language Models (OLLMs) due to the extreme distribution heterogeneity and disparate outlier patterns across modalities. To address this, we propose MorphoQuant, a modality-aware PTQ framework engineered to preserve cross-modal morphology and mitigate outlier loss. Specifically, we introduce Distribution-Aware Bias Compensation (DABC), which selectively absorbs long-tailed outliers into channel-wise biases. This mechanism safeguards outlier magnitudes while maintaining high-precision discretization for dense inliers, thereby preserving accurate discretization across diverse modal distribution. Complementing this, we propose Morphology-Directed Quantization Function Optimization (MDQFO) to co-optimize the quantization grid with the bias mask, ensuring fine-grained alignment across modalities. Extensive evaluations on Qwen2.5-Omni across benchmarks like MMMU and Video-MME demonstrate our approach's superiority. Notably, our W4A4 model achieves 76.63\% on ScienceQA, significantly outperforming SOTA W4A4 methods and surprisingly surpassing the W4A16 baseline, which fully demonstrates the exceptional accuracy-efficiency trade-off of our framework.
\end{abstract}

\section{Introduction}
\label{sec:intro}


Omni-modal Large Language Models (OLLMs)~\cite{qwen2vl,xu2025qwen3,team2024gemini} have recently achieved unprecedented breakthroughs across a myriad of complex reasoning tasks, ranging from visual question answering~\cite{jin2025efficient,ghosh2024exploring,caffagni2024revolution,xu2023multimodal,bhuyan2025bvqa} to embodied AI~\cite{feng2025multi,omnivtla}. By aligning diverse sensory inputs, OLLMs exhibit cognitive paradigms that closely mirror human perception, primarily driven by their large-scale multimodal datasets~\cite{liu2025datasets,li2024survey,yin2023lamm} and massive parameter scales. However, their exorbitant memory demands and computational cost pose severe impediments to real-world deployment, particularly on resource-constrained or edge devices. Furthermore, the auto-regressive nature of OLLM response generation necessitates extensive memory bandwidth for repetitive forward passes, severely bottlenecking inference latency. Consequently, compressing model size and mitigating memory consumption without compromising reasoning capability has emerged as a paramount challenge.

\begin{wrapfigure}{r}{0.5\textwidth} 
    \centering
    \vspace{-10pt} 
    \includegraphics[width=0.48\textwidth]{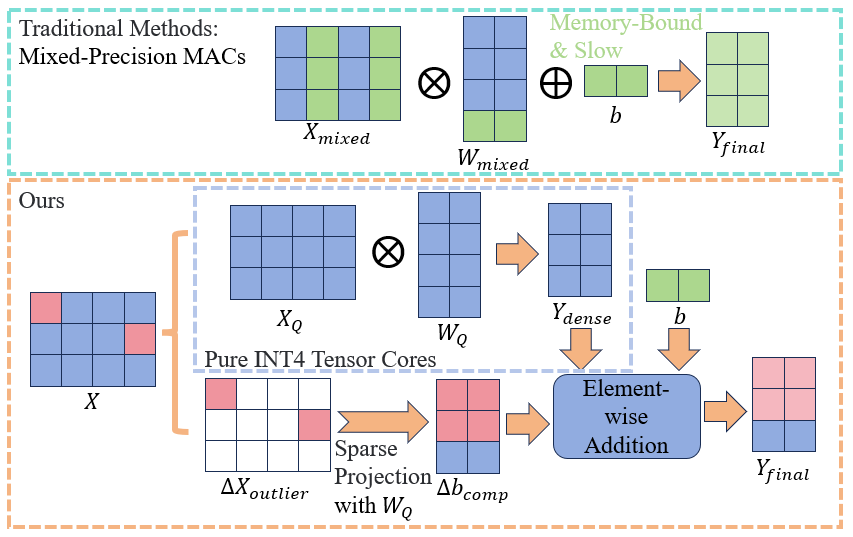} 
    \vspace{-10pt} 
    \captionsetup{margin=10pt, width=0.48\textwidth}
    \caption{\textbf{Comparison with traditional methods.}  \textbf{(Top)} Traditional mixed-precision methods necessitate memory-interleaved operations (blue-INT4, green-FP16), suffering from fragmented memory access and severe overhead. \textbf{(Bottom)} Our DABC strictly decouples the computation, skilfully dividing the activation into the vast majority in pure INT4 Tensor Cores and the sparse outliers (red) undergoing a lightweight projection.}
    \label{Fig.comparison}
    \vspace{-10pt} 
\end{wrapfigure}

To alleviate these computational bottlenecks, various model compression techniques have been explored, including pruning~\cite{pruning_ref}, low-rank decomposition~\cite{lora_ref}, effective architecture design~\cite{bai2024beyond} and quantization~\cite{smoothquant,awq}. Among these, Post-Training Quantization (PTQ)~\cite{qslaw,qvlm,mquant,wang2024towards,wang2024q} stands out as a highly practical paradigm that maps floating-point representations into low-bit integers using a minuscule calibration dataset, which significantly accelerates inference while circumventing the prohibitive costs of full retraining. However, pushing PTQ to the ultra-low W4A4 (4-bit weight, 4-bit activation) extreme for true \textit{omni-modal} models which simultaneously ingest audio, video, images, and text like Qwen2.5-Omni~\cite{xu2025qwen25omnitechnicalreport}, poses a formidable challenge. The inherent dynamics of these diverse sensory inputs introduce highly heterogeneous cross-modal information, giving rise to disparate outlier patterns and severe distribution discrepancies. Existing quantization frameworks~\cite{huang2024empirical} fail to adequately address this cross-modal distribution chasm due to applying uniform quantization functions across these heterogeneous modalities, leading to catastrophic outlier truncation with substantial discretization errors for dense inliers. While some recent methods attempt to suppress these outliers~\cite{smoothquant,llm_int8,spqr}, they still inevitably suffer from severe precision degradation under extreme low bit-widths. Conversely, mixed-precision workarounds~\cite{atom} retain 16-bit matrix multiplications for salient channels, failing to fully exploit the computational efficiency of pure 4-bit quantization. Therefore, discovering a hardware-efficient mechanism to simultaneously tackle disparate outlier patterns and align heterogeneous cross-modal distributions remains a critical, unresolved challenge.

\vspace{-1px}

To bridge this gap, we propose a novel modality-aware PTQ framework, MorphoQuant, engineered for omni-modal large language models which mitigate severe cross-modal distribution discrepancies through targeted outlier compensation and omni-modal distribution alignment directed by morphology. Specifically, we first introduce the Distribution-Aware Bias Compensation (DABC) mechanism to neutralize the impact of disparate outlier patterns. We mathematically formalize a dispersion score to quantify channel-wise eccentricity, enabling the dynamic identification of channels with pronounced outlier tails. Instead of resorting to throughput-disrupting mixed-precision arithmetic, DABC extracts the substantial truncation residuals from these salient channels and seamlessly absorbs them into a channel-wise bias term. This approach effectively safeguards the vital magnitude of outliers while maintaining a high-precision quantization grid for dense inlier distributions. Furthermore, we observe that once these extreme outliers are decoupled, the residual activations exhibit a distinct morphology characterized by a zero-symmetric body with a one-sided long tail. Capitalizing on this trait, we further introduce a Morphology-Directed Quantization Function Optimization (MDQFO) strategy which co-optimizes the discretization mapping and the bias compensation strategy through a collaborative search process. Guided by a redesigned morphology-focused composite loss function, this optimization ensures the precise reconstruction of compact inliers while rigorously preserving fine-grained semantic alignment across all modalities. Consequently, our framework achieves a superior accuracy-efficiency trade-off, unlocking the potential for seamless deployment of omni-modal intelligence on standard 4-bit hardware.

Remarkably, under the extreme W4A4 configuration, our framework not only delivers exceptional reasoning performance but also surprisingly surpasses the SOTA W4A16 baselines on several rigorous benchmarks. This phenomenon unequivocally substantiates the exceptional accuracy-efficiency trade-off enabled by our framework, revealing the profound potential of strategic, modality-aware compensation. Our main contributions are summarized as follows:
\vspace{-8px}
\begin{itemize}
    \item We present the first systematic investigation into the ultra-low bit (W4A4) quantization bottlenecks of all-in-one omni-modal LLMs, successfully addressing a critical gap in current omni-modal compression research.
    \item We propose the Distribution-Aware Bias Compensation (DABC) mechanism, a hardware-efficient approach that neutralizes catastrophic outlier clipping errors by absorbing truncation residuals into a channel-wise bias, thereby circumventing the throughput bottlenecks of mixed-precision arithmetic.
    \item We introduce the Morphology-Directed Quantization Function Optimization (MDQFO) strategy, which co-optimizes the discretization mapping and bias compensation via a novel morphology-focused composite loss, maximizing the semantic fidelity of the compact inlier distribution.
    \item Extensive evaluations on the Qwen2.5-Omni model across a comprehensive suite of omni-modal benchmarks incorporating audio, video, and complex visual reasoning demonstrate that our W4A4 method achieves unprecedented accuracy, astonishingly eclipsing traditional W4A16 approaches.
\end{itemize}

\begin{figure*}[htbp]
\centering
\includegraphics[width=0.9\textwidth]{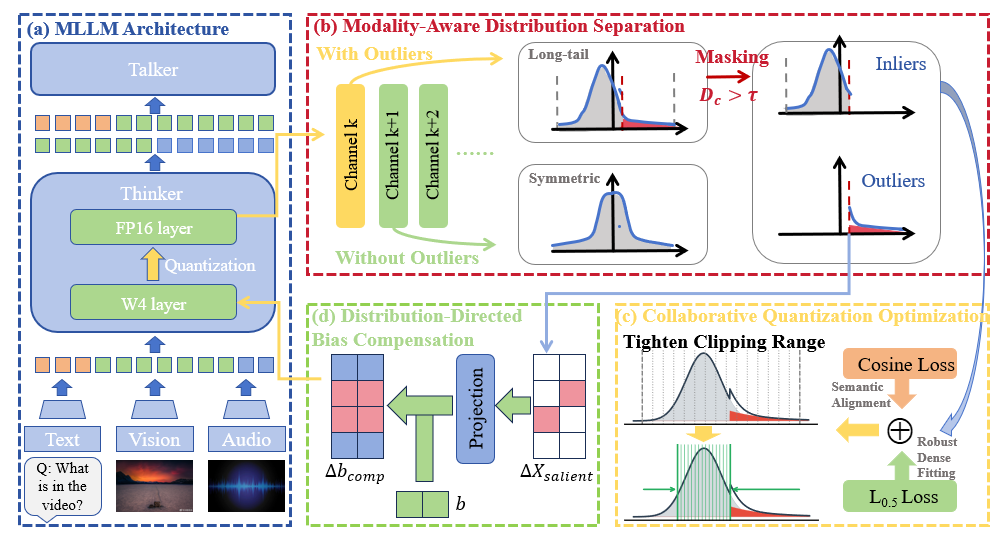}
\vspace{-12pt}
\captionsetup{margin=10pt, width=0.9\textwidth}
\caption{\textbf{Overview of our omni-modal quantization framework.} \textbf{(a) OLLM Architecture}: Omni-modal inputs converge into Qwen2.5-Omni, leading to severe distribution disparity. \textbf{(b) Modality-Aware Distribution Separation}: We identify long-tailed channels via dispersion score $\mathcal{D}_c$, separating activations into symmetric inliers and sparse outliers. \textbf{(d) Distribution-Directed Bias Compensation}: Outliers are projected and merged into the original bias $b$ to form $\Delta b_{comp}$. \textbf{(c) Collaborative Quantization Optimization}: Inliers are quantized with tightened boundaries guided by a composite loss ($\mathcal{L}_{p} \& \mathcal{L}_{cos}$).}
\label{Fig.pipeline}
\end{figure*}

\section{Related Works}
\label{sec:related}

\subsection{Post-Training Quantization of Single-Modal Language Models}

Post-training quantization (PTQ)~\cite{yao2022zeroquant,wu2020easyquant,huang2024billm} of single-modal large language models (LLMs) has been widely studied to reduce memory footprint and inference cost while preserving accuracy. SmoothQuant~\cite{smoothquant} proposes a training-free framework that smooths channel-wise activation outliers into weights via per-channel scaling, enabling hardware-friendly W8A8 quantization on very large LLMs with minimal performance degradation. QLoRA~\cite{qlora} focuses on memory-efficient fine-tuning by storing pretrained weights in a 4-bit NormalFloat (NF4) format, combining double quantization of scales with low-rank adapters trained in higher precision, which allows models with tens of billions of parameters to be finetuned on a single GPU. AWQ~\cite{awq} targets on-device deployment and uses activation statistics to identify a small set of salient channels; it applies equivalent per-channel scaling to protect roughly 0.1--1\% of weights, achieving effective W4 weight-only quantization (W4A16). 

More recent methods introduce learnable transformations to further improve low-bit PTQ. OmniQuant~\cite{omniquant} combines learnable weight clipping (LWC) with learnable equivalent transformations (LET) on activations, optimized in a block-wise manner to calibrate quantization parameters across bit-widths including W4A4. SpinQuant~\cite{spinquant} learns orthogonal rotations on residual streams, attention projections, and KV caches, parameterized on the Stiefel manifold, to redistribute outliers into more isotropic directions and improve W4A4KV4 quantization of text-only LLMs. These approaches largely focus on unimodal transformers and provide the foundation for low-bit compression of large language models.

\subsection{Quantization of Multimodal and Vision--Language Models}

Quantization of multimodal large language models (MLLMs)~\cite{shinde2025survey} and vision--language models (VLMs)~\cite{wang2025bi} has recently attracted increasing interest as these models are deployed in broader application scenarios. QSLAW~\cite{qslaw} studies on quantization-aware adaptation of MLLMs and introduces quantization-aware scale learning with a multimodal warmup schedule, where group-wise weight scales are learned jointly with visual instruction tuning.
For large vision--language models, Q-VLM~\cite{qvlm} proposes a PTQ framework that explicitly considers cross-layer dependency of discretization errors; it employs Cross-layer Dependency Mining (CDM) to group layers into blocks based on activation entropy and Visual Encoder Optimization (VEO) to reshape visual activations by jointly optimizing entropy and reconstruction error. 

At the same time, MBQ (Modality-Balanced Quantization)~\cite{mbq} observes that language and vision tokens exhibit different sensitivities to quantization and introduces gradient-based sensitivity indicators to reweight reconstruction losses for different modalities within channel-wise equalization, along with a mean-absolute-error objective that better correlates with loss change. MQuant~\cite{mquant} further tailors PTQ to MLLMs by identifying challenges in prefill latency, modality-wise distribution gaps, and rotation-induced outliers; it proposes Modality-Specific Static Quantization (MSQ) with distinct static scales for visual and textual tokens, Attention-Invariant Flexible Switching (AIFS) for token reordering under preserved attention, and Rotation Magnitude Suppression (RMS) to handle channels affected by Hadamard-based rotations. These works highlight the importance of modality-aware design when extending PTQ techniques from text-only LLMs to multimodal and vision--language settings.

\subsection{Bias Correction and Outlier Mitigation}

Bias correction and outlier mitigation provide another pathway for improving the robustness of quantized language models. Outlier Suppression+ (OS+)~\cite{osplus} systematically analyzes activation outliers and finds that they are concentrated in a few channels and strongly asymmetric across channels; it introduces channel-wise shifting to re-center activations and channel-wise scaling to compress high-range channels, and shows how these transformations can be equivalently migrated into subsequent weights and biases. OS+ selects outlier thresholds via grid search on calibration data to directly minimize output reconstruction error and is applied to both standard per-tensor and fine-grained quantization schemes on models such as BERT, OPT, BLOOM, and LLaMA. Related approaches, including SmoothQuant and OmniQuant~\cite{smoothquant,omniquant}, also employ equivalent scaling and shifting to reshape activation and weight distributions before quantization within transformer architectures, forming a broad family of methods that explicitly manipulate activation statistics to facilitate low-bit quantization.
\section{Approach}
\label{sec:approach}

In this section, we present our post-training quantization framework tailored for OLLMs. We begin by reviewing standard PTQ in Section~\ref{subsec:preliminaries} and explicitly identifying the cross-modal distribution chasm in Section~\ref{subsec:motivation}. To address this, we introduce the Distribution-Aware Bias Compensation (DABC) in Section~\ref{subsec:bias_comp} to precisely protect essential outliers with negligible latency overhead, followed by a Morphology-Directed Quantization Function Optimization (MDQFO) in Section~\ref{subsec:threshold_design} to maximize the representational fidelity of the inlier distribution.

\subsection{Preliminaries: Uniform Quantization and Its Limitations}
\label{subsec:preliminaries}

For a standard linear quantization process, given a full-precision activation tensor $\mathbf{X}$ and weight matrix $\mathbf{W}$, the uniform quantization maps the continuous values into a $k$-bit integer grid. The quantized activation $\mathbf{X}_Q$ and weight $\mathbf{W}_Q$ are formulated as:
\begin{equation}
    \mathbf{X}_Q = \text{clip}\left(\left\lfloor \frac{\mathbf{X}}{s_x} \right\rceil, Q_{min}, Q_{max}\right), \quad \mathbf{W}_Q = \text{clip}\left(\left\lfloor \frac{\mathbf{W}}{s_w} \right\rceil, Q_{min}, Q_{max}\right),
    \label{eq:standard_quant}
\end{equation}
where $s_x$ and $s_w$ denote the quantization scaling factors, and $\lfloor \cdot \rceil$ represents the round-to-nearest operator and $[Q_{min}, Q_{max}]$ defines the representational boundaries. The fundamental quantization error $\Delta \mathbf{X}$ consists of two mutually exclusive components: \textit{clipping error} $\mathcal{E}_{clip}$ and \textit{rounding error} $\mathcal{E}_{round}$:
\begin{equation}
    \Delta \mathbf{X} = \underbrace{(\mathbf{X}_{out} - \alpha_c)}_{\mathcal{E}_{clip}: \text{ Clipping error}} + \underbrace{(\mathbf{X}_{in} - s_x \mathbf{X}_Q)}_{\mathcal{E}_{round}: \text{ Rounding error}}.
    \label{eq:error_decomposition}
\end{equation}
Here, $\mathbf{X}_{out}$ and $\mathbf{X}_{in}$ respectfully represent outliers and inliers in the activation, and $\alpha_c$ is the clipping threshold derived from the scaling factor $s_x$. 
The objective of conventional PTQ is typically formulated to minimize the output reconstruction error with $\ell_2$-norm:
\begin{equation}
    \min_{s_x, s_w} \left\| \mathbf{W}\mathbf{X} - (s_w \mathbf{W}_Q)(s_x \mathbf{X}_Q) \right\|_2^2
    \label{eq:ptq_objective}.
\end{equation}
Specifically, this $\ell_2$-norm objective is inherently sensitive to the \textbf{disparate outlier patterns and distribution characteristics} prevalent in omni-modal models, which precipitates a critical dilemma for selection of quantization threshold $\alpha_c$ under extreme low-bit constraints. Expanding the quantization range to accommodate outliers inflates the step size, thereby amplifying rounding errors $\mathcal{E}_{\text{round}}$ for dense inliers; conversely, tightening the range to preserve inlier precision results in severe clipping errors $\mathcal{E}_{\text{clip}}$ for critical outliers. Both scenarios substantially increase the total quantization error $\Delta \mathbf{X}$.

\subsection{Motivation: The Cross-Modal Distribution Chasm}
\label{subsec:motivation}

\begin{figure}[htbp]
     \centering
     \begin{subfigure}[b]{0.35\textwidth}
         \centering
         \includegraphics[width=\textwidth]{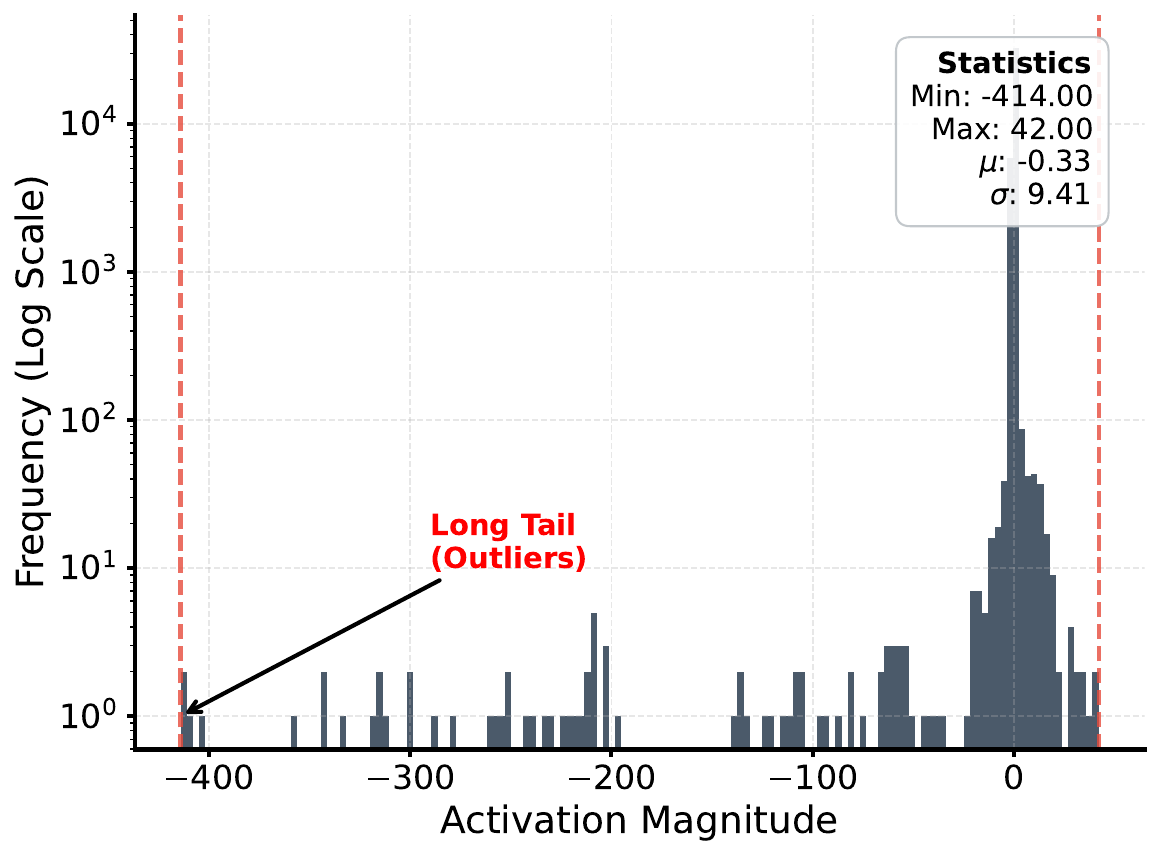}
         \caption{Activation distribution}
         \label{fig:histogram}
     \end{subfigure}
     \hfill 
     \begin{subfigure}[b]{0.27\textwidth}
         \centering
         \includegraphics[width=\textwidth]{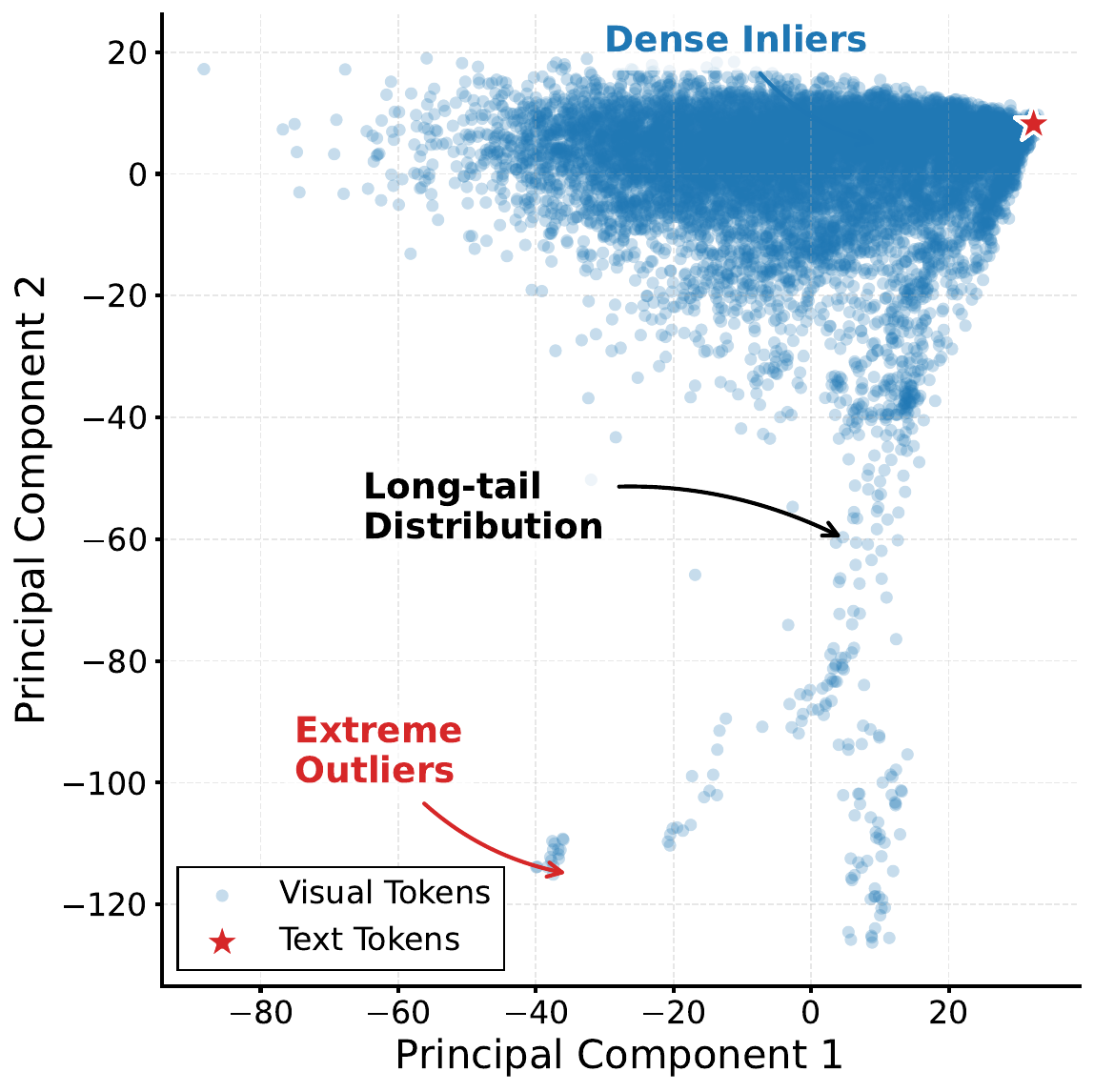}
         \caption{PCA of activation}
         \label{fig:modality_dist}
     \end{subfigure}
     \hfill 
     \begin{subfigure}[b]{0.32\textwidth}
         \centering
         \includegraphics[width=\textwidth]{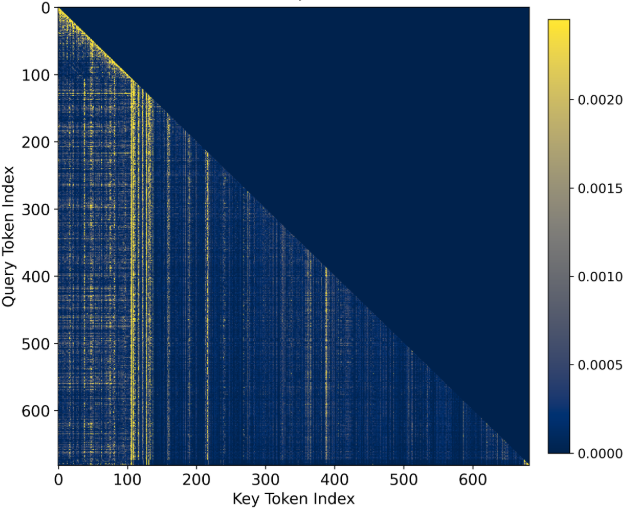}
         \caption{Attention map error}
         \label{fig:attn_map}
     \end{subfigure}
     
     \caption{\textbf{(Left)} A layer with extreme long tail, while the inliers dominate a tiny range. \textbf{(Middle)} PCA of activation inputs (Video-MME), indicating cross-modal disparity, long tail and sparse outliers. \textbf{(Right)} Quantization error of an attention map, in which the primary error lies along the modal boundary line.}
     \label{fig:three_graphs}
\end{figure}

To elucidate the root causes of the threshold selection dilemma established in Section~\ref{subsec:preliminaries}, we conduct a comprehensive empirical analysis of the activation distributions within omni-modal architectures. Unlike conventional Vision-Language Models (VLMs)~\cite{qslaw,qvlm,mquant} that process bimodal inputs, omni-modal systems integrate heterogeneous sensory data into a unified backbone:
\begin{equation}
    \mathbf{X}_{omni} = \text{Concat}\left( \text{Proj}_m(\text{Enc}_m(\mathbf{I}_m)) \right), \quad m \in \{\text{Audio, Video, Image, Text}\}.
    \label{eq:omni_concat}
\end{equation}
This architectural convergence leads to a catastrophic representation collapse under extreme low-bit quantization. Specifically, our empirical analysis reveals three critical phenomena:
\begin{itemize}
    \item \textbf{Significant Outlier Tails:} As visualized in activation histograms (Fig.~\ref{fig:histogram}), specific channels exhibit extreme outlier values that dominate the dynamic range, complicating optimal threshold selection.
    \item \textbf{Cross-Modal Morphological Disparity:} Dissecting the unified distribution into isolated modalities (Fig.~\ref{fig:modality_dist}) reveals fundamentally disparate magnitudes and outlier morphologies. This forces a scaling dilemma: global uniform scaling factors either severely clip crucial high-magnitude features or drown subtle features in rounding noise.
    \item \textbf{Semantic Boundary Degradation:} Attention map visualizations (Fig.~\ref{fig:attn_map}) corroborate that prominent quantization degradation occurs precisely at semantic boundaries between modalities, disrupting cross-modal reasoning.
\end{itemize}

Based on these observation, we conclude that conventional methods appear substantial discretization errors for omni-modal models under extreme low-bitwidths, as the uniform quantization functions fail to accommodate both the heavy tails and the dense inlier bodies simultaneously. Thus, the pivotal challenge is: \textit{How can we precisely protect these vital modality-specific outliers while maintaining fine-grained preservation of normal activations (inliers) to prevent cross-modal semantic loss?}

\subsection{Distribution-Aware Bias Compensation}
\label{subsec:bias_comp}


To address the pivotal challenge identified in Section~\ref{subsec:motivation}, we propose the Distribution-Aware Bias Compensation (DABC) mechanism. The core principle is to strategically shift the clipping loss of activations into a channel-wise bias compensation term. This preserves critical outlier information while maintaining accurate quantization for the majority of inliers, without sacrificing hardware efficiency.

Specifically, we first evaluate the dispersion metric for each channel to identify the minuscule fraction of activations demanding protection. The dispersion score $\mathcal{D}_c$ of the input $\mathbf{X}_c$ in channel $c$ is formulated as:
\begin{equation}
    \mathcal{D}_c = \frac{\left| \max(\mathbf{X}_c) - \min(\mathbf{X}_c) \right|}{\|\overline{\mathbf{X}_c}\|_1 + \epsilon},
    \label{eq:dispersion}
\end{equation}
where $\epsilon$ ensures numerical stability. $\left| \max(\mathbf{X}_c) - \min(\mathbf{X}_c) \right|$ measures the range of data distribution, and outliers can cause this value to increase significantly. The denominator normalizes by the mean absolute value $\frac{1}{N} \|\mathbf{X}_c\|_1$, representing the density of the inlier body. 

Once the dispersion metric for all channels is computed, we sort them in descending order and employ an adjustable threshold $\tau$ to partition the channels to strike an optimal accuracy-efficiency trade-off. For the salient channels marked by $\mathcal{D}_c > \tau$, their severe truncation residuals are shielded from the destructive $k$-bit grid and instead rerouted into the bias compensation term. Specifically, we generate a channel-wise binary mask $\mathbf{m} \in \{0, 1\}^{C_{in}}$ to pinpoint the unbalanced channels with long tails:
\begin{equation}
    m_c = \mathbb{I}(\mathcal{D}_c > \tau),
    \label{eq:mask}
\end{equation}
where $\mathbb{I}(\cdot)$ denotes the indicator function. Instead of discarding the catastrophic clipping errors induced by these heavy-tailed channels, we extract their truncated outlier residuals, defined as $\Delta \mathbf{X}_{salient} = (\mathbf{X} - \mathbf{X}_Q) \odot \mathbf{m}$, where $\odot$ represents the broadcasted element-wise multiplication. This critical outlier information is then projected through the quantized weights $\mathbf{W}_Q$ and mathematically folded into the bias term. The fully compensated forward pass is thus formulated as:
\begin{equation}
    \mathbf{Y} = (\mathbf{W}_{Q} \mathbf{X}_{Q}) + \mathbf{b} + \underbrace{\mathbf{W}_Q \left( (\mathbf{X} - \mathbf{X}_Q) \odot \mathbf{m} \right)}_{\Delta \mathbf{b}_{comp}}.
    \label{eq:bias_comp}
\end{equation}
By decoupling heavy-tailed activations, DABC safeguards the vital magnitude of outliers within the high-precision bias term $\Delta\mathbf{b}_{\text{comp}}$, while maintaining high-precision discretization for dense inlier distributions. 

\noindent\textbf{Compatibility Advantages over Mixed-Precision.} 
It is imperative to distinguish DABC from existing mixed-precision outlier protection paradigms, such as ATOM~\cite{atom}. Methods like ATOM retain FP16 values for salient channels, forcing the hardware to execute mixed-precision Matrix Multiply-Accumulate operations (MACs), which severely disrupts the dense computation throughput of Tensor Cores. In stark contrast, DABC strictly maintains a \textit{pure 4-bit dense matrix multiplication} ($\mathbf{W}_{Q} \otimes \mathbf{X}_{Q}$) for the compute-intensive phase, and introduces the high-precision outlier correction purely as a lightweight 16-bit sparse matrix multiplication and element-wise bias addition. Consequently, this architectural decoupling makes DABC intrinsically highly compatible with highly-optimized 4-bit CUDA kernels, guaranteeing seamless hardware deployment and true inference acceleration without compromising precision.

\begin{figure}
\centering
\vspace{-8pt}
\includegraphics[width=0.6\textwidth]{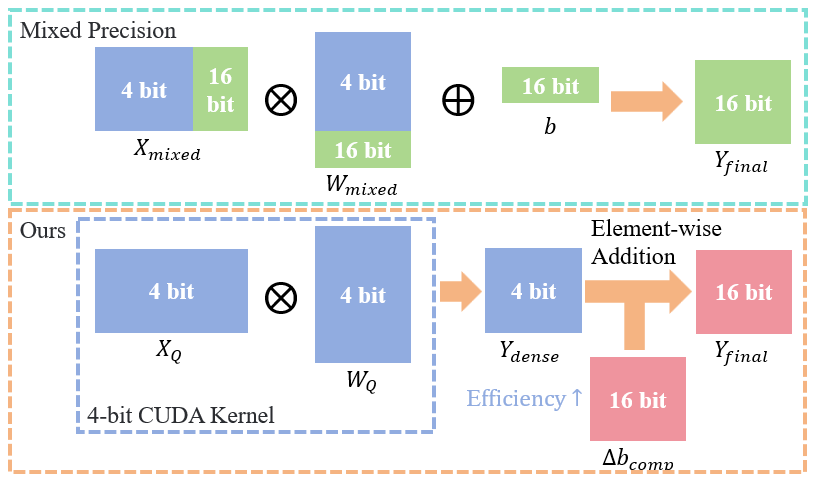}
\vspace{-6pt}
\caption{\textbf{Hardware-friendly Execution Flow.} \textbf{(Top)} Traditional mixed-precision methods suffer from memory-interleaved operations. \textbf{(Bottom)} Our approach executes the vast majority of activations via standard dense 4-bit CUDA kernels, while integrating sparse outlier corrections through lightweight element-wise addition, ensuring hardware-friendly inference.}
\label{Fig.calculation}
\vspace{-8pt}
\end{figure}

\subsection{Morphology-Directed Quantization Function Optimization}
\label{subsec:threshold_design}

Building on the compact, zero-symmetric distribution established by DABC, we propose the Morphology-Directed Quantization Function Optimization (MDQFO) strategy to refine the quantization grid. Following the previous section, data with outliers removed exhibits a highly regular zero-symmetric distribution, enabling us to aggressively tighten the clipping boundary $\alpha_c$ optimization strategy based on the intrinsic morphological traits of the activation data: \textit{zero-symmetric inlier} and \textit{long-tailed outlier}.




Given zero-symmetry, we strictly restrict the k-bit quantization range of the compliant channels to a symmetric interval $[-\alpha_c, \alpha_c]$, where the clipping boundary $\alpha_c$ acts as a learnable parameter. 
To optimally reconstruct the residual distribution, we discard the conventional Mean Squared Error (MSE), which is overly sensitive to residual noise. Instead, we employ an $\ell_{p}$-norm loss, which places greater emphasis on accurately reconstructing the dense inliers:
\begin{equation}
    \mathcal{L}_{p} = \frac{1}{N} \sum_{i=1}^{N} \sqrt{\left| \mathbf{X}_{FP}^{(i)} - \hat{\mathbf{X}}_Q^{(i)}(\alpha_c) \right|},
    \label{eq:l05_loss}
\end{equation}
where $\hat{\mathbf{X}}_Q = s_x \mathbf{X}_Q + \Delta \mathbf{X}_{salient}$ is the dequantized value together with compensation, and $N$ is the total number of elements. Furthermore, as the directional alignment of high-dimensional feature vectors is crucial for cross-modal semantic interaction in OLLMs, we introduce a Cosine Similarity Loss $\mathcal{L}_{cos}$ to strictly preserve the spatial angle between the full-precision and quantized activations:
\begin{equation}
    \mathcal{L}_{cos} = 1 - \frac{1}{M} \sum_{j=1}^{M} \frac{\mathbf{X}_{FP}^{(j)} \cdot \hat{\mathbf{X}}_Q^{(j)}(\alpha_c)}{\left\| \mathbf{X}_{FP}^{(j)} \right\|_2 \left\| \hat{\mathbf{X}}_Q^{(j)}(\alpha_c) \right\|_2},
    \label{eq:cos_loss}
\end{equation}
where $M$ denotes the sequence length, and $\mathbf{X}^{(j)}$ represents the feature vector of the $j$-th token. Then, the overall optimization objective is formulated as:
\begin{equation}
    \min_{\alpha_c} \mathcal{L}_{total} = \mathcal{L}_{p} + \lambda _{cos}\mathcal{L}_{cos},
    \label{eq:joint_search}
\end{equation}
where $\lambda_{cos}$ is a balancing hyperparameter to scale the two loss components. we jointly optimize the learnable clipping boundary $\alpha_c$ by minimizing a composite loss function $\mathcal{L}_{total}$.

Under this collaborative search framework, the gradient of the threshold $\alpha_c$ is not merely driven by the standalone quantization error, but dynamically guided by the presence of DABC. By iteratively updating $\alpha_c$, the algorithm automatically converges to an elegant equilibrium: the Bias Compensation exclusively secures the minuscule fraction of extreme spikes, while the tightened clipping threshold $\alpha_c$ meticulously captures the fine-grained semantics of the vast majority of zero-symmetric inliers. This morphology-directed Divide-and-Conquer strategy systematically maximizes the semantic fidelity across the omni-modal feature space.

\section{Experiments}
\label{sec:experiments}

In this section, we comprehensively evaluate our proposed quantization framework on the Qwen2.5-Omni model. We first detail the experimental setup, encompassing the omni-modal benchmarks and implementation details. Subsequently, we compare our approach against state-of-the-art (SOTA) multimodal quantization methods to demonstrate its superiority under extreme low-bit configurations. Finally, we provide an extensive ablation study to validate the individual contributions of our core algorithmic designs.

\subsection{Experimental Setup}
\label{subsec:exp_setup}

\noindent\textbf{Models and Benchmarks.} 
We deploy our Post-Training Quantization (PTQ) framework on the pre-trained Qwen2.5-Omni (3B) model. To rigorously validate the robustness of our method across the entire omni-modal spectrum, we deliberately depart from the conventional practice of evaluating solely on vision-language tasks. Instead, our evaluation suite spans highly diverse modalities: \textbf{ScienceQA} and \textbf{MMMU} for complex vision-language reasoning, \textbf{Video-MME} for long-context video comprehension, and \textbf{AIR-Bench} for dense audio signal understanding. This holistic benchmark selection ensures our quantized model is evaluated on all modalities inherently supported by MLLMs.

\noindent\textbf{Baselines for Comparison.} 
We benchmark our framework against leading quantization methodologies. To fully demonstrate the efficacy of our approach on extremely low-bit activations, we evaluate both W4A16 and W4A4 configurations. For the W4A16 setting, we compare with widely adopted baselines including AWQ~\cite{awq} and QLoRA~\cite{qlora}. For the extreme W4A4 setting, we compare against recent multimodal quantization techniques such as Q-VLM~\cite{qvlm} and representative mixed-precision methods. We also provide the full-precision (FP16) results derived from the official pre-trained model and codebase as the theoretical upper bound (Oracle).

\noindent\textbf{Implementation Details.} 
To ensure a strictly fair comparison and isolate the performance gains yielded by our activation compression, all W4A4 evaluated methods employ the identical 4-bit weight compression technique, adopted from QLoRA, for the MLLM backbone. Our Distribution-Aware Bias Compensation (DABC) and Morphology-Directed Quantization Function Optimization are applied exclusively to the activation tensors. During the calibration and search phase, we sample a minuscule calibration set comprising 128 multimodal samples. The dispersion threshold $\tau$ is empirically set to mask the top \text{5\%} channels.

\subsection{Comparison with State-of-the-Art Methods}
\label{subsec:main_results}

To the best of our knowledge, we are the pioneers in systematically pushing the quantization of true omni-modal large language models to the W4A4 extreme. By conducting evaluations across text, image, video, and audio modalities, we bridge a critical void in current MLLM quantization research, which has predominantly been restricted to bi-modal (image-text) architectures.

\begin{table}[t]
\centering
\resizebox{\textwidth}{!}{
\begin{tabular}{lccccccc}
\toprule
\multirow{2}{*}{Method} & \multirow{2}{*}{W/A Bits} & \multicolumn{2}{c}{Vision-Language} & Video & Audio & \multirow{2}{*}{Memory (GB)} \\ \cmidrule(lr){3-4} \cmidrule(lr){5-5} \cmidrule(lr){6-6}
 &  & ScienceQA & MMMU & Video-MME & AIR-Bench &  \\ \midrule
FP16 (Full Precision) & 16/16 & 79.39 & 44.78 & 57.33 & 66.25 & 11.79 \\ \midrule
\rowcolor{gray!10} \multicolumn{7}{l}{\textit{State-of-the-art PTQ Baselines}} \\
QLoRA & 4/16 & 75.88 & 43.56 & \textbf{54.78} & 64.12 & 5.54 \\ \midrule
\rowcolor{gray!10} \multicolumn{7}{l}{\textit{Ultra-low bit (W4A4) Quantization}} \\
Naive W4A4 & 4/4 & 73.17 & 40.00 & 47.52 & 63.40 & 6.02 \\
QVLM & 4/4 & 73.85 & 40.33 & 48.41 & 63.71 & 5.03 \\
PoMQ-ViT & 4/4 & 75.36 & 42.11 & 49.22 & 64.17 & 6.37 \\
\textbf{Ours} & \textbf{4/4} & \textbf{76.63}$^{\dagger}$ & \textbf{45.11}$^{\dagger}$ & \textbf{54.33} & \textbf{65.09}$^{\dagger}$ & 6.07 \\ \bottomrule
\end{tabular}%
}
\caption{Main results of Qwen2.5-Omni quantization across various multimodal benchmarks. We report accuracy (\%) for ScienceQA, MMMU, and Video-MME, and the overall score for AIR-Bench. Memory represents the GPU VRAM consumption of model weights and active buffers on ScienceQA. \textbf{Bold} indicates the best results in the 4-bit category. $\dagger$ denotes that our W4A4 method surpasses the W4A16 baseline.}
\label{tab:main_results}
\vspace{-15pt}
\end{table}
\begin{table}[htbp]
\centering
\resizebox{\textwidth}{!}{%
\begin{tabular}{l c cccccccc}
\toprule
\multirow{2}{*}{\textbf{Method}} & \multirow{2}{*}{\textbf{W/A Bits}} & \multicolumn{2}{c}{\textbf{Short (\%)}} & \multicolumn{2}{c}{\textbf{Medium (\%)}} & \multicolumn{2}{c}{\textbf{Long (\%)}} & \multicolumn{2}{c}{\textbf{Overall (\%)}} \\ 
\cmidrule(lr){3-4} \cmidrule(lr){5-6} \cmidrule(lr){7-8} \cmidrule(lr){9-10}
 &  & w/o subs & w/ subs & w/o subs & w/ subs & w/o subs & w/ subs & w/o subs & w/ subs \\ \midrule
 
\multicolumn{10}{c}{\textit{Full Precision Reference}} \\ \midrule
FP16 (Oracle) & 16/16 & 70.83 & 70.75 & 58.97 & 52.69 & 41.67 & 47.60 & 65.38 & 55.65 \\ \midrule

\multicolumn{10}{c}{\textit{W4A16 Baselines}} \\ \midrule
QLoRA & 4/16 & 69.10 & 66.67 & 54.49 & 51.34 & 54.17 & 44.75 & 63.46 & 52.96 \\ \midrule

\multicolumn{10}{c}{\textit{Ultra-low Bit (W4A4) Methods}} \\ \midrule
Naive W4A4 & 4/4 & 63.54 & 57.52 & 45.51 & 44.22 & 29.10 & 38.93 & 55.77 & 45.79 \\
QVLM & 4/4 & 63.89 & 57.35 & 42.95 & 47.31 & 37.50 & 39.27 & 55.56 & 46.91 \\
PoMQ-ViT & 4/4 & 64.24 & 58.17 & 52.56 & 45.83 & 45.83 & 40.41 & 59.40 & 47.09 \\
\textbf{Ours (Full)} & 4/4 & \textbf{69.79} & \textbf{67.32} & \textbf{51.92} & \textbf{49.87} & \textbf{45.83} & \textbf{44.63} & \textbf{62.61} & \textbf{52.60} \\ \bottomrule
\end{tabular}%
}
\caption{Detailed performance breakdown on the \textbf{Video-MME} benchmark. We evaluate our quantized Qwen2.5-Omni model across different video lengths (Short, Medium, Long) and subtitle conditions (w/o subs, w/ subs). \textbf{Bold} indicates the best performance among 4-bit quantization methods.}
\label{tab:video_mme_detailed}
\vspace{-15pt}
\end{table}

\noindent\textbf{Performance on Omni-modal Benchmarks.}
As summarized in Table~\ref{tab:main_results}, our proposed framework demonstrates superior robustness across all evaluated modalities, consistently outperforming existing W4A4 quantization methods by a significant margin.

\noindent\textbf{Surpassing W4A16 Baselines.}
The most striking observation is that our W4A4 method achieves accuracy comparable to, or even exceeding, the W4A16 QLoRA baseline. Specifically, on \textbf{ScienceQA}, our method reaches \textbf{76.63\%}, surpassing QLoRA (75.88\%) by 0.75\%. On the challenging visual-text \textbf{MMMU} benchmark, we achieve \textbf{45.11\%}, which not only outperforms QLoRA (43.56\%) but surprisingly exhibits a slight improvement over the FP16 Oracle (44.78\%). We attribute this phenomenon to the regularization effect of our morphology-directed quantization, which effectively filters out noise in the visual features while preserving semantic alignment. Similarly, on the audio-centric \textbf{AIR-Bench}, our method (65.09\%) outperforms QLoRA (64.12\%), demonstrating that our DABC mechanism effectively handles high-frequency outliers in audio modalities better than simply retaining 16-bit activations.

\noindent\textbf{Robustness in Video Comprehension.}
Video understanding typically suffers the most from quantization due to temporal feature accumulation. While Naive W4A4 and QVLM degrade significantly on \textbf{Video-MME} (47.52\% and 48.41\%, respectively), our method maintains a high accuracy of \textbf{54.33\%}, dramatically narrowing the gap with the W4A16 baseline (54.78\%). This validates that our collaborative threshold design successfully captures the temporal dynamics in long-context video streams, ensuring that the critical motion semantics are not lost in the 4-bit bottleneck.

\begin{figure}[htbp]
    \centering
    \captionsetup[subfigure]{aboveskip=2pt, belowskip=2pt}
    
    \begin{subfigure}[b]{0.32\textwidth}
        \centering
        \includegraphics[width=\textwidth]{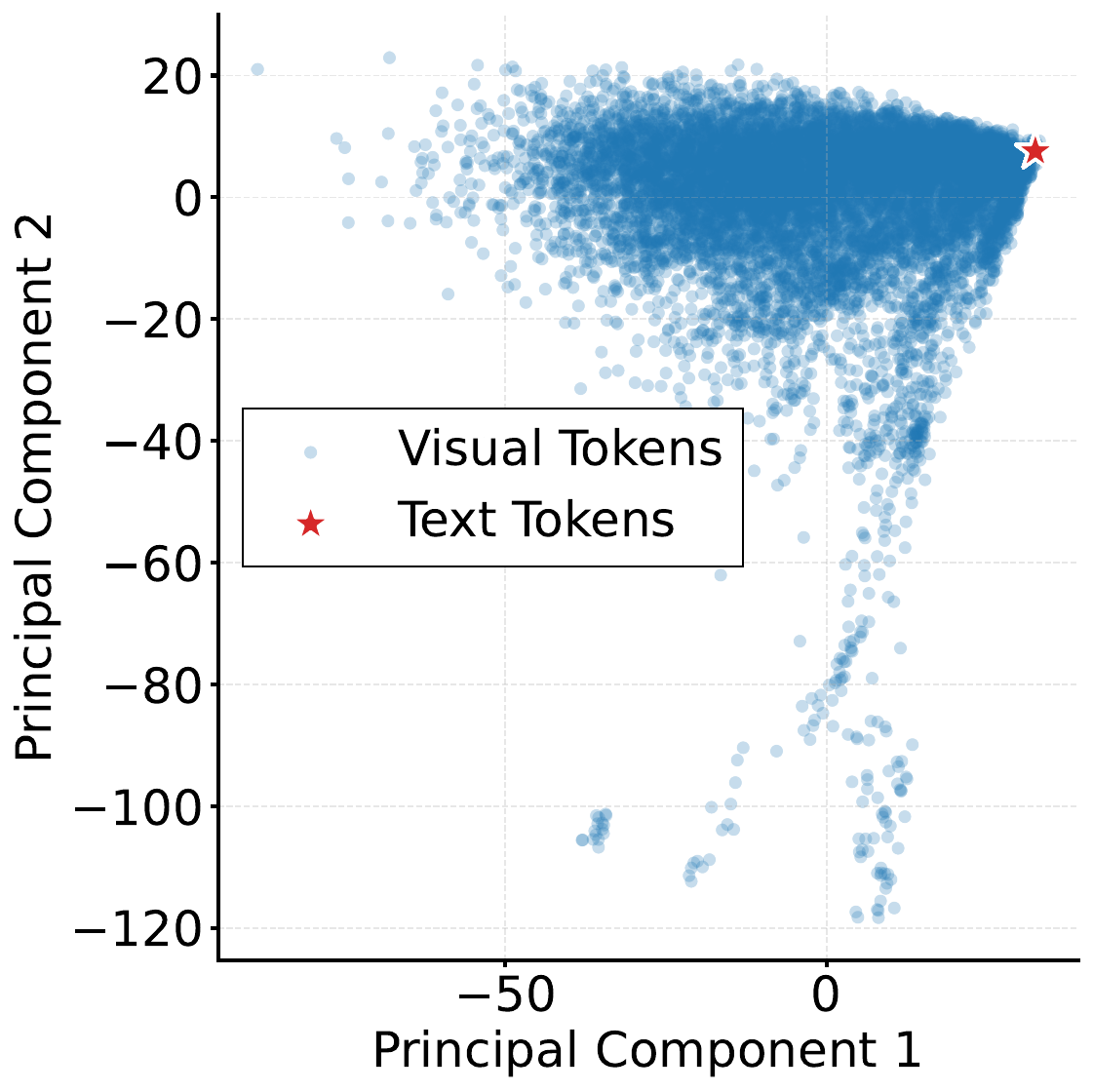}
        \caption{FP16 (Oracle)}
        \label{fig:vis_fp16}
    \end{subfigure}
    \hfill
    \begin{subfigure}[b]{0.32\textwidth}
        \centering
        \includegraphics[width=\textwidth]{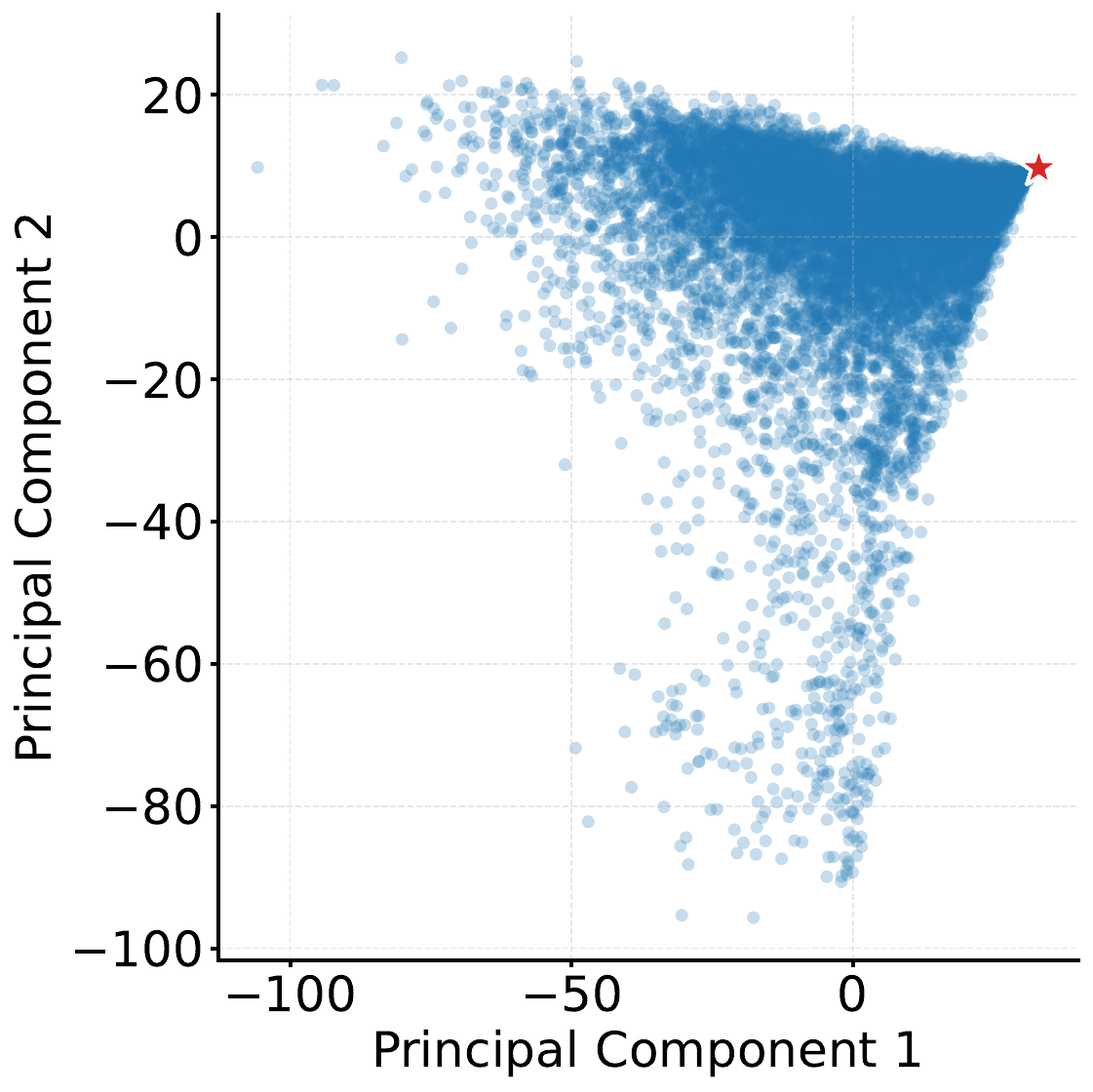}
        \caption{Q-VLM(W4A4)}
        \label{fig:vis_qvlm}
    \end{subfigure}
    \hfill
    \begin{subfigure}[b]{0.32\textwidth}
        \centering
        \includegraphics[width=\textwidth]{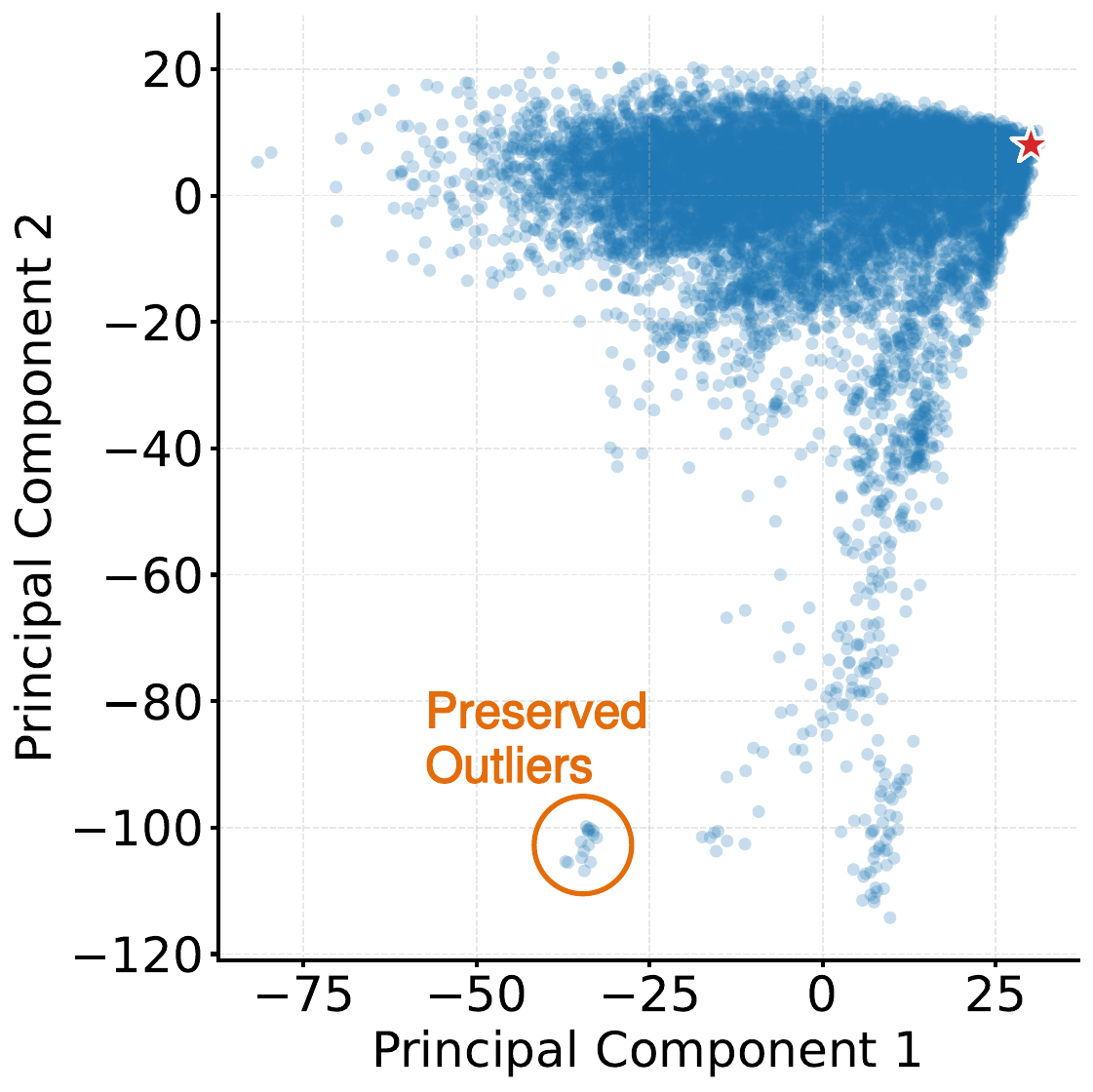}
        \caption{\textbf{Ours(W4A4)}}
        \label{fig:vis_ours}
    \end{subfigure}
    
    \vspace{-1ex}
    \caption{\textbf{PCA visualization of activation features (Thinker Layer 0).} Compared to (a) the FP16 oracle, (b) traditional W4A4 quantization like Q-VLM suffers from severe \textbf{representation collapse}, losing the significant outliers. Conversely, (c) \textbf{our method} faithfully restores the original feature manifold and preserved outliers, preserving cross-modal semantic alignment.}
    \label{fig:qualitative_comparison}
    
\end{figure}

\noindent\textbf{Maintaining Omni-Modal Semantic.}
The superiority in preserving the semantics of our framework is validated through the PCA visualization of the feature space in Fig.~\ref{fig:qualitative_comparison}. As observed in the FP16 oracle Fig.~\ref{fig:vis_fp16}, omni-modal activations exhibit a distinctive feature with long tail, extreme sparse outliers and dense symmetric inliers, which is essential for distinguishing cross-modal semantics. While methods like Q-VLM causes the feature morphology to collapse in Fig.~\ref{fig:vis_qvlm}, our mechanism retained the original structure in Fig.~\ref{fig:vis_ours}. This evidence suggests that \textit{preserving the distribution morphology} is more critical for reasoning fidelity than increasing the overall bit-width.

\subsection{Ablation Study}
\label{subsec:ablation}


To thoroughly validate the efficacy of each proposed component, we conduct a progressive ablation study on the ScienceQA and MMMU benchmarks under the W4A4 setting, as detailed in Table~\ref{tab:ablation}.

\noindent\textbf{Impact of Core Components.} 
Table~\ref{tab:ablation} dissects the progressive performance gains. Starting from the baseline, W4A4 Q-VLM, the model suffers from significant quantization noise, achieving only 73.85\% on ScienceQA.
\begin{itemize}
    \item \textbf{Effectiveness of DABC \& Collaborative Search:} By integrating our proposed Distribution-Aware Bias Compensation (DABC) and the associated Morphology-Directed Quantization Function Optimization, the accuracy surges to \textbf{75.81\%}. This substantial gain (+1.96\%) confirms that isolating long-tailed outliers and tightening the inlier grid are the primary drivers for recovering model capability.
    \item \textbf{Necessity of Composite Loss:} Introducing our tailored objective function further boosts the performance to \textbf{76.63\%}. This leap allows our W4A4 model to remarkably surpass the W4A16 QLoRA reference (75.88\%), validating that semantic alignment by introducing $\mathcal{L}_{cos}$ and robust dense fitting via $\mathcal{L}_{p}$ are crucial for fine-tuning the quantization boundaries.
\end{itemize}

\begin{table}[t]
\centering
\resizebox{\textwidth}{!}{
\begin{tabular}{ccccccc}
\toprule
\multirow{2}{*}{Variant} & \multirow{2}{*}{DABC} & \multirow{2}{*}{Collab. Search} & \multirow{2}{*}{Comp. Loss} & \multicolumn{2}{c}{Accuracy (\%)} \\ \cmidrule(lr){5-6}
 & & & & ScienceQA & MMMU \\ \midrule
Baseline & - & - & - & 73.85 & 40.33 \\
+ DABC & \checkmark & \checkmark & - & 75.81 & 44.22 \\
Ours (Full) & \checkmark & \checkmark & \checkmark & \textbf{76.63} & \textbf{45.11} \\ \midrule
\textit{Reference} & \multicolumn{3}{c}{\textit{QLoRA (W4A16)}} & \textit{75.88} & \textit{43.56} \\
\bottomrule
\end{tabular}%
}
\caption{\textbf{Ablation study on Qwen2.5-Omni (W4A4)}. We progressively verify the effectiveness of DABC, Collaborative Search, and Composite Loss. Remarkably, our full pipeline outperforms the W4A16 QLoRA baseline on ScienceQA and MMMU.}
\label{tab:ablation}
\vspace{-15pt}
\end{table}

\begin{figure}[htbp]
     \centering
     \begin{subfigure}[b]{0.45\textwidth}
         \centering
         \includegraphics[width=\textwidth]{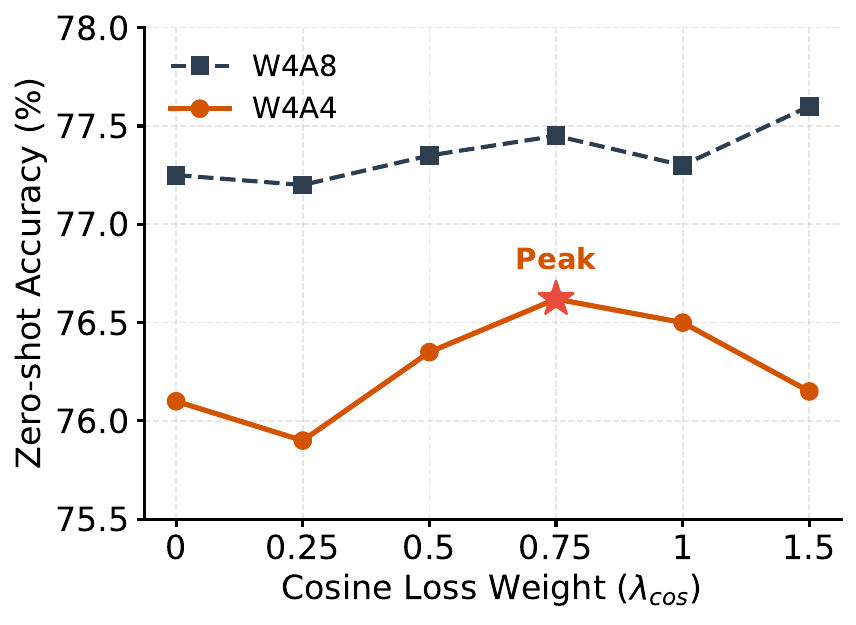}
         \caption{Accuracy on different $\lambda_{cos}$}
         \label{fig:ablation_1}
     \end{subfigure}
     \hfill 
     \begin{subfigure}[b]{0.45\textwidth}
         \centering
         \includegraphics[width=\textwidth]{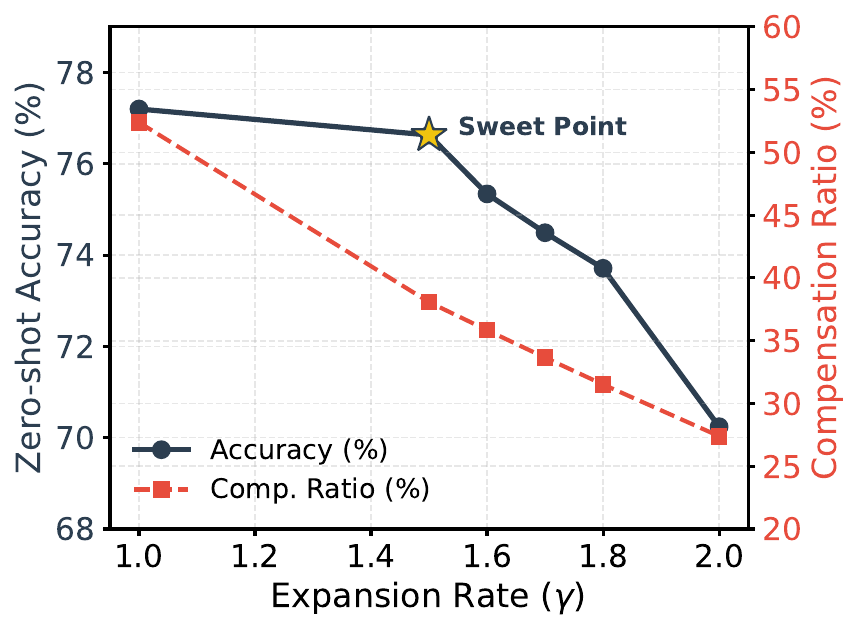}
         \caption{Performance on different $\gamma$}
         \label{fig:ablation_2}
     \end{subfigure}
     
     \caption{\textbf{Ablation study on hyper-parameters.} \textbf{(Left)} The answering accuracy w.r.t. different $\lambda_
         {cos}$ on ScienceQA dataset. \textbf{(Right)} The accuracy and compensation percentage w.r.t. different expansion rate on ScienceQA dataset.}
     \label{fig:ablation_curves}
\vspace{-10px}
\end{figure}

\noindent\textbf{Sensitivity to Hyperparameters.} 
We further investigate the robustness of our framework against key hyperparameters in Fig.~\ref{fig:ablation_curves}.
\begin{itemize}
    \item \textbf{Balance of Loss Components ($\lambda_{cos}$):} As shown in Fig.~\ref{fig:ablation_curves}(a), the model performance is sensitive to the weight of the Cosine Loss at 4-bits. The accuracy peaks at $\lambda_{cos}=0.75$. Lower values ($\lambda_{cos} < 0.5$) fail to enforce semantic direction alignment, while excessive values ($\lambda_{cos} > 1.0$) dominate the optimization, hampering the magnitude reconstruction of dense inliers. At 8-bits, the limited variation in accuracy is likely due to the inherently high precision of the measurement itself.
    \item \textbf{Trade-off: Accuracy vs. Compensation Overhead:} Fig.~\ref{fig:ablation_curves}(b) analyzes the effect of the expansion rate $\gamma$, which acts as a multiplicative factor to the learned quantization threshold $\alpha_c$. Activations are compensated only when their magnitude exceeds $\gamma \cdot \alpha_c$. 
    \textbf{Analysis:} As the expansion rate increases from 1.0 to 2.0, the \textbf{compensation ratio} decreases at the cost of accuracy degradation dropping from $\sim$77\% to $\sim$70\%. $\gamma=1.5$ is the sweet point to cover the long-tail outliers for optimal performance.
\end{itemize}

\section{Conclusion}
\label{sec:conclusion}



In this paper, we proposed a novel PTQ framework specifically tailored for MLLMs, with successful deployment and rigorous evaluation on the Qwen2.5-Omni architecture. Departing from conventional outlier suppression paradigms, we introduced Distribution-Aware Bias Compensation (DABC), which strategically folds long-tailed outliers into bias terms based on channel dispersion score. This guaranties high inlier precision without incurring mixed-precision overhead. Furthermore, our Morphology-Directed Quantization Function Optimization (MDQFO) jointly optimizes quantization boundaries with the bias mask using a customized composite loss to preserve cross-modal alignment. Extensive evaluations demonstrate that our framework significantly outperforms SOTA W4A4 methods, remarkably achieving accuracy comparable to or even surpassing W4A16 baselines across diverse omni-modal benchmarks, demonstrating the powerful potential of modality-aware quantization optimization strategy. Future work will continue to explore more advanced, fine-grained and hardware-efficient outlier processing methodologies to further close the gap with full-precision omni-modal intelligence.
\nocite{yang2022lavt}


\clearpage

\bibliographystyle{splncs}
\bibliography{egbib}

@article{qvlm,
  title={Q-vlm: Post-training quantization for large vision-language models},
  author={Wang, Changyuan and Wang, Ziwei and Xu, Xiuwei and Tang, Yansong and Zhou, Jie and Lu, Jiwen},
  journal={Advances in Neural Information Processing Systems},
  volume={37},
  pages={114553--114573},
  year={2024}
}

@inproceedings{smoothquant,
  title={Smoothquant: Accurate and efficient post-training quantization for large language models},
  author={Xiao, Guangxuan and Lin, Ji and Seznec, Mickael and Wu, Hao and Demouth, Julien and Han, Song},
  booktitle={International conference on machine learning},
  pages={38087--38099},
  year={2023},
  organization={PMLR}
}

@article{qlora,
  title={Qlora: Efficient finetuning of quantized llms},
  author={Dettmers, Tim and Pagnoni, Artidoro and Holtzman, Ari and Zettlemoyer, Luke},
  journal={Advances in neural information processing systems},
  volume={36},
  pages={10088--10115},
  year={2023}
}

@article{awq,
  title={Awq: Activation-aware weight quantization for on-device llm compression and acceleration},
  author={Lin, Ji and Tang, Jiaming and Tang, Haotian and Yang, Shang and Chen, Wei-Ming and Wang, Wei-Chen and Xiao, Guangxuan and Dang, Xingyu and Gan, Chuang and Han, Song},
  journal={Proceedings of machine learning and systems},
  volume={6},
  pages={87--100},
  year={2024}
}

@article{omniquant,
  title={Omniquant: Omnidirectionally calibrated quantization for large language models},
  author={Shao, Wenqi and Chen, Mengzhao and Zhang, Zhaoyang and Xu, Peng and Zhao, Lirui and Li, Zhiqian and Zhang, Kaipeng and Gao, Peng and Qiao, Yu and Luo, Ping},
  journal={arXiv preprint arXiv:2308.13137},
  year={2023}
}

@article{spinquant,
  title={Spinquant: Llm quantization with learned rotations},
  author={Liu, Zechun and Zhao, Changsheng and Fedorov, Igor and Soran, Bilge and Choudhary, Dhruv and Krishnamoorthi, Raghuraman and Chandra, Vikas and Tian, Yuandong and Blankevoort, Tijmen},
  journal={arXiv preprint arXiv:2405.16406},
  year={2024}
}

@inproceedings{qslaw,
  title={Advancing multimodal large language models with quantization-aware scale learning for efficient adaptation},
  author={Xie, Jingjing and Zhang, Yuxin and Lin, Mingbao and Cao, Liujuan and Ji, Rongrong},
  booktitle={Proceedings of the 32nd ACM International Conference on Multimedia},
  pages={10582--10591},
  year={2024}
}

@inproceedings{mbq,
  title={Mbq: Modality-balanced quantization for large vision-language models},
  author={Li, Shiyao and Hu, Yingchun and Ning, Xuefei and Liu, Xihui and Hong, Ke and Jia, Xiaotao and Li, Xiuhong and Yan, Yaqi and Ran, Pei and Dai, Guohao and others},
  booktitle={Proceedings of the Computer Vision and Pattern Recognition Conference},
  pages={4167--4177},
  year={2025}
}

@inproceedings{mquant,
  title={Mquant: Unleashing the inference potential of multimodal large language models via static quantization},
  author={Yu, JiangYong and Zhou, Sifan and Yang, Dawei and Li, Shuoyu and Wang, Shuo and Hu, Xing and Xu, Chen and Xu, Zukang and Shu, Changyong and Yuan, Zhihang},
  booktitle={Proceedings of the 33rd ACM International Conference on Multimedia},
  pages={1783--1792},
  year={2025}
}

@inproceedings{osplus,
  title={Outlier suppression+: Accurate quantization of large language models by equivalent and effective shifting and scaling},
  author={Wei, Xiuying and Zhang, Yunchen and Li, Yuhang and Zhang, Xiangguo and Gong, Ruihao and Guo, Jinyang and Liu, Xianglong},
  booktitle={Proceedings of the 2023 Conference on Empirical Methods in Natural Language Processing},
  pages={1648--1665},
  year={2023}
}

@article{spqr,
  title={Spqr: A sparse-quantized representation for near-lossless llm weight compression},
  author={Dettmers, Tim and Svirschevski, Ruslan and Egiazarian, Vage and Kuznedelev, Denis and Frantar, Elias and Ashkboos, Saleh and Borzunov, Alexander and Hoefler, Torsten and Alistarh, Dan},
  journal={arXiv preprint arXiv:2306.03078},
  year={2023}
}

@article{atom,
  title={Atom: Low-bit quantization for efficient and accurate llm serving},
  author={Zhao, Yilong and Lin, Chien-Yu and Zhu, Kan and Ye, Zihao and Chen, Lequn and Zheng, Size and Ceze, Luis and Krishnamurthy, Arvind and Chen, Tianqi and Kasikci, Baris},
  journal={Proceedings of Machine Learning and Systems},
  volume={6},
  pages={196--209},
  year={2024}
}

@article{qwen2vl,
  title={Qwen2-vl: Enhancing vision-language model's perception of the world at any resolution},
  author={Wang, Peng and Bai, Shuai and Tan, Sinan and Wang, Shijie and Fan, Zhihao and Bai, Jinze and Chen, Keqin and Liu, Xuejing and Wang, Jialin and Ge, Wenbin and others},
  journal={arXiv preprint arXiv:2409.12191},
  year={2024}
}

@article{omnivtla,
  title={Omnivtla: Vision-tactile-language-action model with semantic-aligned tactile sensing},
  author={Cheng, Zhengxue and Zhang, Yiqian and Zhang, Wenkang and Li, Haoyu and Wang, Keyu and Song, Li and Zhang, Hengdi},
  journal={arXiv preprint arXiv:2508.08706},
  year={2025}
}

@article{jin2025efficient,
  title={Efficient multimodal large language models: A survey},
  author={Jin, Yizhang and Li, Jian and Gu, Tianjun and Liu, Yexin and Zhao, Bo and Lai, Jinxiang and Gan, Zhenye and Wang, Yabiao and Wang, Chengjie and Tan, Xin and others},
  journal={Visual Intelligence},
  volume={3},
  number={1},
  pages={27},
  year={2025},
  publisher={Springer}
}

@article{xu2023multimodal,
  title={Multimodal learning with transformers: A survey},
  author={Xu, Peng and Zhu, Xiatian and Clifton, David A},
  journal={IEEE Transactions on Pattern Analysis and Machine Intelligence},
  volume={45},
  number={10},
  pages={12113--12132},
  year={2023},
  publisher={Ieee}
}

@article{ghosh2024exploring,
  title={Exploring the frontier of vision-language models: A survey of current methodologies and future directions},
  author={Ghosh, Akash and Acharya, Arkadeep and Saha, Sriparna and Jain, Vinija and Chadha, Aman},
  journal={arXiv preprint arXiv:2404.07214},
  year={2024}
}

@article{caffagni2024revolution,
  title={The revolution of multimodal large language models: A survey},
  author={Caffagni, Davide and Cocchi, Federico and Barsellotti, Luca and Moratelli, Nicholas and Sarto, Sara and Baraldi, Lorenzo and Cornia, Marcella and Cucchiara, Rita},
  journal={Findings of the association for computational linguistics: ACL 2024},
  pages={13590--13618},
  year={2024}
}

@article{bhuyan2025bvqa,
  title={BVQA: connecting language and vision through multimodal attention for open-ended question answering},
  author={Bhuyan, Md Shalha Mucha and Hossain, Eftekhar and Sathi, Khaleda Akhter and Hossain, Md Azad and Dewan, M Ali Akber},
  journal={IEEE Access},
  year={2025},
  publisher={IEEE}
}

@article{feng2025multi,
  title={Multi-agent embodied ai: Advances and future directions},
  author={Feng, Zhaohan and Xue, Ruiqi and Yuan, Lei and Yu, Yang and Ding, Ning and Liu, Meiqin and Gao, Bingzhao and Sun, Jian and Zheng, Xinhu and Wang, Gang},
  journal={arXiv preprint arXiv:2505.05108},
  year={2025}
}

@article{liu2025datasets,
  title={Datasets for large language models: A comprehensive survey},
  author={Liu, Yang and Cao, Jiahuan and Liu, Chongyu and Ding, Kai and Jin, Lianwen},
  journal={Artificial Intelligence Review},
  volume={58},
  number={12},
  pages={403},
  year={2025},
  publisher={Springer}
}

@article{li2024survey,
  title={A survey on benchmarks of multimodal large language models},
  author={Li, Jian and Lu, Weiheng and Fei, Hao and Luo, Meng and Dai, Ming and Xia, Min and Jin, Yizhang and Gan, Zhenye and Qi, Ding and Fu, Chaoyou and others},
  journal={arXiv preprint arXiv:2408.08632},
  year={2024}
}

@article{yin2023lamm,
  title={Lamm: Language-assisted multi-modal instruction-tuning dataset, framework, and benchmark},
  author={Yin, Zhenfei and Wang, Jiong and Cao, Jianjian and Shi, Zhelun and Liu, Dingning and Li, Mukai and Huang, Xiaoshui and Wang, Zhiyong and Sheng, Lu and Bai, Lei and others},
  journal={Advances in Neural Information Processing Systems},
  volume={36},
  pages={26650--26685},
  year={2023}
}

@inproceedings{pruning_ref,
  title={Fit and prune: Fast and training-free visual token pruning for multi-modal large language models},
  author={Ye, Weihao and Wu, Qiong and Lin, Wenhao and Zhou, Yiyi},
  booktitle={Proceedings of the AAAI Conference on Artificial Intelligence},
  volume={39},
  number={21},
  pages={22128--22136},
  year={2025}
}

@article{lora_ref,
  title={Llava-mole: Sparse mixture of lora experts for mitigating data conflicts in instruction finetuning mllms},
  author={Chen, Shaoxiang and Jie, Zequn and Ma, Lin},
  journal={arXiv preprint arXiv:2401.16160},
  year={2024}
}

@article{huang2024empirical,
  title={An empirical study of llama3 quantization: From llms to mllms},
  author={Huang, Wei and Zheng, Xingyu and Ma, Xudong and Qin, Haotong and Lv, Chengtao and Chen, Hong and Luo, Jie and Qi, Xiaojuan and Liu, Xianglong and Magno, Michele},
  journal={Visual Intelligence},
  volume={2},
  number={1},
  pages={36},
  year={2024},
  publisher={Springer}
}

@article{llm_int8,
  title={Gpt3. int8 (): 8-bit matrix multiplication for transformers at scale},
  author={Dettmers, Tim and Lewis, Mike and Belkada, Younes and Zettlemoyer, Luke},
  journal={Advances in neural information processing systems},
  volume={35},
  pages={30318--30332},
  year={2022}
}

@article{yao2022zeroquant,
  title={Zeroquant: Efficient and affordable post-training quantization for large-scale transformers},
  author={Yao, Zhewei and Yazdani Aminabadi, Reza and Zhang, Minjia and Wu, Xiaoxia and Li, Conglong and He, Yuxiong},
  journal={Advances in neural information processing systems},
  volume={35},
  pages={27168--27183},
  year={2022}
}

@article{wu2020easyquant,
  title={Easyquant: Post-training quantization via scale optimization},
  author={Wu, Di and Tang, Qi and Zhao, Yongle and Zhang, Ming and Fu, Ying and Zhang, Debing},
  journal={arXiv preprint arXiv:2006.16669},
  year={2020}
}

@article{huang2024billm,
  title={Billm: Pushing the limit of post-training quantization for llms},
  author={Huang, Wei and Liu, Yangdong and Qin, Haotong and Li, Ying and Zhang, Shiming and Liu, Xianglong and Magno, Michele and Qi, Xiaojuan},
  journal={arXiv preprint arXiv:2402.04291},
  year={2024}
}

@article{shinde2025survey,
  title={A Survey on Efficient Vision-Language Models},
  author={Shinde, Gaurav and Ravi, Anuradha and Dey, Emon and Sakib, Shadman and Rampure, Milind and Roy, Nirmalya},
  journal={Wiley Interdisciplinary Reviews: Data Mining and Knowledge Discovery},
  volume={15},
  number={3},
  pages={e70036},
  year={2025},
  publisher={Wiley Online Library}
}

@article{wang2025bi,
  title={Bi-vlm: Pushing ultra-low precision post-training quantization boundaries in vision-language models},
  author={Wang, Xijun and Huang, Junyun and Abdalla, Rayyan and Zhang, Chengyuan and Xian, Ruiqi and Manocha, Dinesh},
  journal={arXiv preprint arXiv:2509.18763},
  year={2025}
}

@misc{xu2025qwen25omnitechnicalreport,
      title={Qwen2.5-Omni Technical Report}, 
      author={Jin Xu and Zhifang Guo and Jinzheng He and Hangrui Hu and Ting He and Shuai Bai and Keqin Chen and Jialin Wang and Yang Fan and Kai Dang and Bin Zhang and Xiong Wang and Yunfei Chu and Junyang Lin},
      year={2025},
      eprint={2503.20215},
      archivePrefix={arXiv},
      primaryClass={cs.CL},
      url={https://arxiv.org/abs/2503.20215}, 
}

@article{bai2024beyond,
  title={Beyond efficiency: A systematic survey of resource-efficient large language models},
  author={Bai, Guangji and Chai, Zheng and Ling, Chen and Wang, Shiyu and Lu, Jiaying and Zhang, Nan and Shi, Tingwei and Yu, Ziyang and Zhu, Mengdan and Zhang, Yifei and others},
  journal={arXiv preprint arXiv:2401.00625},
  year={2024}
}

@article{xu2025qwen3,
  title={Qwen3-omni technical report},
  author={Xu, Jin and Guo, Zhifang and Hu, Hangrui and Chu, Yunfei and Wang, Xiong and He, Jinzheng and Wang, Yuxuan and Shi, Xian and He, Ting and Zhu, Xinfa and others},
  journal={arXiv preprint arXiv:2509.17765},
  year={2025}
}

@article{team2024gemini,
  title={Gemini 1.5: Unlocking multimodal understanding across millions of tokens of context},
  author={Team, Gemini and Georgiev, Petko and Lei, Ving Ian and Burnell, Ryan and Bai, Libin and Gulati, Anmol and Tanzer, Garrett and Vincent, Damien and Pan, Zhufeng and Wang, Shibo and others},
  journal={arXiv preprint arXiv:2403.05530},
  year={2024}
}

@inproceedings{yang2022lavt,
  title={Lavt: Language-aware vision transformer for referring image segmentation},
  author={Yang, Zhao and Wang, Jiaqi and Tang, Yansong and Chen, Kai and Zhao, Hengshuang and Torr, Philip HS},
  booktitle={Proceedings of the IEEE/CVF conference on computer vision and pattern recognition},
  pages={18155--18165},
  year={2022}
}

@inproceedings{wang2024towards,
  title={Towards accurate post-training quantization for diffusion models},
  author={Wang, Changyuan and Wang, Ziwei and Xu, Xiuwei and Tang, Yansong and Zhou, Jie and Lu, Jiwen},
  booktitle={Proceedings of the IEEE/CVF Conference on Computer Vision and Pattern Recognition},
  pages={16026--16035},
  year={2024}
}

@article{wang2024q,
  title={Q-vlm: Post-training quantization for large vision-language models},
  author={Wang, Changyuan and Wang, Ziwei and Xu, Xiuwei and Tang, Yansong and Zhou, Jie and Lu, Jiwen},
  journal={Advances in Neural Information Processing Systems},
  volume={37},
  pages={114553--114573},
  year={2024}
}
\end{document}